\documentclass[11pt]{article}

\usepackage[numbers]{natbib}
\usepackage[margin=1in]{geometry}

\usepackage[utf8]{inputenc}
\usepackage[T1]{fontenc}
\usepackage{amsmath,amssymb,amsfonts}
\usepackage{amsthm}
\usepackage{mathtools}
\usepackage{booktabs}
\usepackage{graphicx}
\usepackage{caption}
\usepackage{subcaption}
\usepackage{float}
\usepackage{hyperref}
\usepackage{url}
\usepackage{nicefrac}
\usepackage{enumitem}
\usepackage{microtype}
\usepackage{xcolor}
\graphicspath{{figures/}}

\theoremstyle{definition}
\newtheorem{definition}{Definition}
\newtheorem{assumption}{Assumption}

\theoremstyle{plain}
\newtheorem{theorem}{Theorem}
\newtheorem{corollary}{Corollary}

\usepackage{xcolor}
\usepackage{authblk}

\title{Measurement Geometry and Design for\\ Trustworthy Generative Inverse Problems}

\author[1]{Pengfei Jin \textsuperscript{*}}
\author[2]{Na Li}
\author[1]{Quanzheng Li\textsuperscript{\dag}}

\affil[1]{Center for Advanced Medical Computing and Analysis, Massachusetts General Hospital and Harvard Medical School, Boston, MA 02114}
\affil[2]{School of Engineering and Applied Sciences, Harvard University, Boston, MA 02138}
\date{}

\begin{document}

\maketitle

\renewcommand{\thefootnote}{\fnsymbol{footnote}}
\footnotetext[1]{First author.}
\footnotetext[2]{Corresponding author. Email: \texttt{li.quanzheng@mgh.harvard.edu}.}
\renewcommand{\thefootnote}{\arabic{footnote}}

\begin{abstract}
Generative models are increasingly used as priors for inverse problems, but their ability to produce realistic images creates a basic trust problem: a plausible reconstruction may be supported by the measurements, or it may be filled in by the prior along unobserved directions. This distinction is especially important in medical imaging, where acquisition operators are designed under scan-time, dose, and calibration constraints. We study generative inverse problems from a measurement-geometry perspective. The central question is whether a fixed measurement operator can distinguish nearby images that are plausible under the generative prior, and whether this relationship can guide better measurements. We introduce a local measurement-manifold compatibility measure that quantifies how well the operator observes prior-relevant tangent directions. Under local regularity assumptions, we prove that this quantity controls the stable part of the reconstruction error, while the generative prior controls off-manifold drift. This worst-direction certificate motivates practical fixed and sequential acquisition rules based on overall local volume preservation, including a posterior-cloud design that adapts measurements at test time without training a sampling policy. Across row-sampling, tomographic, and MR acquisition settings, the proposed scores predict failure modes, explain measurement-induced hallucinations, and guide better sampling. In fastMRI Cartesian sampling, posterior-cloud measurement design improves over strong non-learned ACS-preserving baselines, including variable-density and Poisson-like masks.
\end{abstract}

\section{Introduction}

Generative models, especially diffusion models, have become standard priors for image restoration and inverse problems \citep{ho2020denoising,song2021score,dhariwal2021diffusion}. Given measurements
\begin{equation}
y = A x_* + e,
\label{eq:intro-forward-model}
\end{equation}
a diffusion inverse solver can combine a pretrained score or denoiser with data consistency to reconstruct images even when $A$ is highly underdetermined \citep{chung2023dps,song2023pigdm,kawar2022ddrm,wang2022ddnm,zhu2023diffpir,dou2024filtering,daras2024survey}. Medical imaging is a natural target because MRI and CT are constrained by acquisition time, radiation dose, motion, hardware, and patient burden \citep{lustig2007sparse,ye2019csmri,cheng2019compressed,hyun2021deep}. It also exposes the central risk: a plausible-looking output may be a measurement-supported reconstruction, or it may be a hallucination \citep{kim2025tackling,tianrods} along directions that the operator does not observe.

\begin{figure}[t]
\centering
\includegraphics[width=0.7\linewidth]{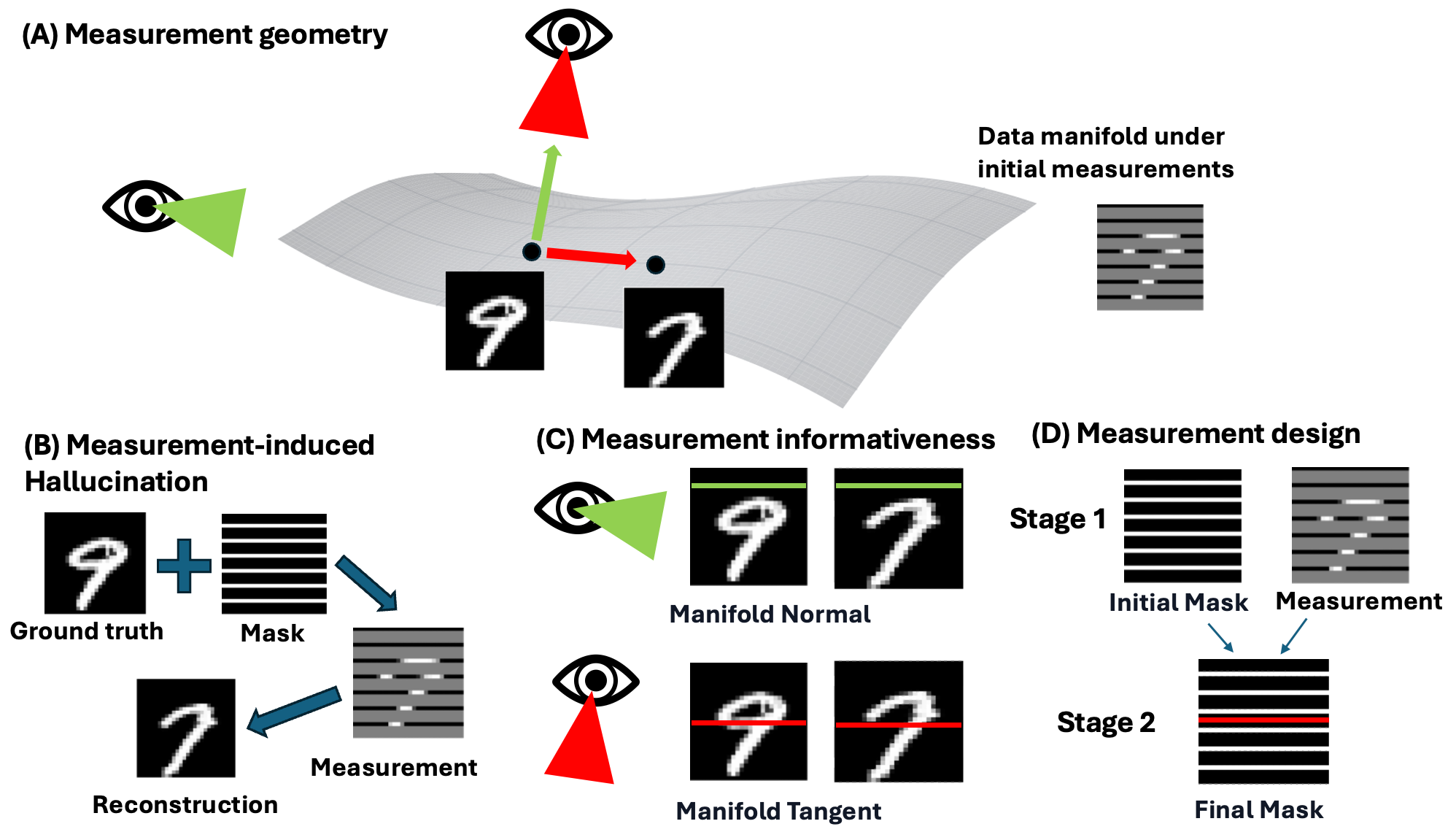}
\caption{Overview of measurement geometry and design. (A) A generative prior induces a data manifold; under the initial measurements, different plausible points on this manifold can remain hard to distinguish. The red direction denotes tangent ambiguity between plausible images, while the green direction denotes a manifold-normal measurement that is less useful for resolving such ambiguity. (B) In row-sampling MNIST, an insufficient mask can make a digit 9 measurement-consistent with a plausible 7, producing a measurement-induced hallucination. (C) Not all feasible measurements are equally informative: a manifold-normal row observes structure shared by the alternatives, whereas a manifold-tangent row crosses a stroke region that separates them. (D) Our two-stage design first uses an initial structured mask to obtain measurements and a posterior cloud, then adds a feasible tangent-informative measurement to reduce the remaining manifold ambiguity.}
\label{fig:overview}
\end{figure}

Figure~\ref{fig:overview} previews our viewpoint. Trustworthy reconstruction depends not only on the strength of the generative prior, but also on whether the measurements distinguish the target from nearby prior-plausible alternatives. Classical compressed sensing formalizes this operator-prior compatibility for sparse, wavelet, total-variation, and low-rank models through incoherence, restricted isometry, and related restricted-geometry conditions \citep{donoho2006compressed,candes2006robust,lustig2007sparse,ye2019csmri}. This viewpoint also shaped practical MRI sampling, where variable-density masks, central calibration regions, and Poisson-disc-style designs preserve low-frequency information while creating incoherent aliasing \citep{lustig2007sparse,dwork2021fast,cheng2019compressed}. Generative priors change the prior model, but not the identifiability problem.

Existing theory for generative compressed sensing studies recovery over generator ranges and score-based priors \citep{bora2017generative,daskalakis2020constant,jalal2020robust,jalal2021mri,nguyen2022provable}, but most guarantees use random or fixed measurement operators with global properties. They do not directly answer the scan-specific acquisition question: after a coarse first measurement, which additional structured measurements best disambiguate the current posterior geometry? We study two related questions:
\begin{enumerate}[leftmargin=*]
\item For a given operator $A$, can the relationship between $A$ and the local data manifold explain whether a generative reconstruction is feasible and trustworthy?
\item If that relationship can be measured, can it be used to design a better operator, including a sample-specific second-stage acquisition?
\end{enumerate}

Our analysis treats a pretrained generative prior as defining a local set of plausible alternatives around the target. A measurement operator is trustworthy only to the extent that it separates motion within this local set. If the operator is insensitive to a tangent direction on the data manifold, two visually plausible images can agree with the same measurements; if it observes that direction, the ambiguity is reduced. We formalize this idea later through a local compatibility index and a stability result that separates tangent ambiguity from normal drift.

The same index also suggests a design principle. In practice, medical acquisition operators are structured: CT chooses projection angles, and Cartesian MRI chooses frequency-encoding lines rather than arbitrary dense matrices. We therefore optimize within feasible operator families. Our designs use local neighborhoods or posterior clouds to estimate which plausible alternatives remain after an initial acquisition, then select additional structured measurements that best reduce the remaining manifold ambiguity. The theory gives a conservative worst-direction view of identifiability, while the design objective improves overall coverage of the local posterior geometry.

The experiments follow this logic across three structured acquisition settings. Row sampling makes measurement-induced hallucination directly visible, tomographic angle selection tests the same design principle for a Radon operator, and MRI Cartesian sampling evaluates calibration-aware k-space line selection. Across these settings, geometry scores track reconstruction behavior and posterior-cloud design improves structured sampling under fixed budgets.

Our contributions are:
\begin{enumerate}[leftmargin=*]
\item We establish a measurement-geometry perspective for trustworthy generative inverse problems. This viewpoint explains hallucination as a failure of the operator to distinguish locally plausible alternatives, and we introduce a local compatibility index that quantifies this risk.
\item We use the index as a principle for optimizing structured measurement operators. The resulting fixed and adaptive rules select feasible measurements from local neighborhoods or test-time posterior clouds, without training a sampling policy.
\item We validate the framework across row sampling, tomographic angle selection, and FastMRI-domain Cartesian sampling. The experiments show that the geometry predicts reconstruction behavior and that posterior-cloud design improves over strong non-learned baselines.
\end{enumerate}

\section{Related Work}
\label{sec:related-work}

\paragraph{Generative models for inverse problems.}
Diffusion inverse solvers combine pretrained priors with data consistency using likelihood, posterior, or manifold guidance \citep{chung2023dps,chung2022mcg}, operator-aware pseudoinverse, spectral, or null-space updates \citep{song2023pigdm,kawar2022ddrm,wang2022ddnm}, and variational, filtering, or plug-and-play formulations \citep{mardani2024variational,zhu2023diffpir,dou2024filtering}. Medical variants include score-based inverse problems, accelerated MRI, 3D reconstruction from 2D priors, and training from corrupted measurements \citep{song2021medical,chung2022score,chung2023solving,aali2024ambient}, while surveys emphasize posterior sampling and data consistency as fidelity mechanisms \citep{daras2024survey}. We are complementary to sampler design: for a fixed solver, we ask whether $A$ separates locally plausible alternatives induced by the prior.

\paragraph{Compressed sensing and generative-prior theory.}
Classical compressed sensing recovers sparse or compressible signals when $A$ satisfies incoherence, restricted isometry, or related restricted-geometry conditions \citep{donoho2006compressed,candes2006robust}; imaging and MRI versions use wavelet/TV priors, incoherent aliasing, variable-density k-space sampling, and central low-frequency preservation \citep{lustig2007sparse,ye2019csmri,cheng2019compressed}. Recent medical-imaging theory phrases solvability through low-dimensional manifolds \citep{hyun2021deep}, and generative compressed sensing replaces sparsity with generator ranges or learned priors, including Langevin and score-based extensions \citep{bora2017generative,daskalakis2020constant,jalal2020robust,jalal2021mri,nguyen2022provable}. The shared requirement is that $A$ be well conditioned on the low-complexity signal set; our distinction is locality, since we study conditioning on the current tangent space rather than random measurements or global properties of $A$.

\paragraph{Medical image sampling operators.}
Medical acquisition already treats operator design as part of reconstruction: SENSE, GRAPPA, and ESPIRiT use coil sensitivity and calibration information, motivating fully sampled calibration or ACS regions \citep{pruessmann1999sense,griswold2002grappa,uecker2014espirit}. Compressed-sensing MRI popularized variable-density sampling, low-frequency preservation, and Poisson-disc-style dispersion \citep{lustig2007sparse,ye2019csmri,dwork2021fast,cheng2019compressed}, and FastMRI-style benchmarks standardize acceleration factors, Cartesian masks, and calibration-aware comparisons \citep{zbontar2018fastmri}. Adaptive and learned acquisition methods use training examples, reconstruction-aware criteria, or joint acquisition-reconstruction optimization \citep{ravishankar2011adaptive,gozcu2018learning,bahadir2020loupe}; our rule instead is scan-specific, using the current scan's posterior cloud, while our MR baselines remain strong non-learned ACS-preserving masks.

\section{Problem Setup}
\label{sec:problem-setup}

We formalize the measurement-geometry viewpoint using the linear observation model
\[
y = A x_* + e,
\]
where $x_* \in \mathbb{R}^n$ is the unknown signal, $A \in \mathbb{R}^{m \times n}$ is the measurement operator, and $e$ is measurement noise. We use the following variational template to cover generator-, score-, denoiser-, and diffusion-induced priors:
\begin{equation}
\hat{x} \in \arg\min_{x \in \mathbb{R}^n}
\Phi_{A,y,\lambda,\theta}(x)
:=
\frac{1}{2}\|Ax-y\|_2^2
\;+\;
\lambda \mathcal{R}_{\theta}(x),
\label{eq:master-objective}
\end{equation}
where $\mathcal{R}_{\theta}$ is any prior term that favors the data manifold induced by the pretrained generative model. The theorem below only uses this local manifold-aware behavior, not the global details of a particular sampler or parameterization.

Let $\mathcal{M}$ denote the prior-relevant local manifold around $x_*$. When $\mathcal{M}$ is differentiable at $x_*$, we write
\[
T_* := T_{x_*}\mathcal{M},
\qquad
U_* \in \mathbb{R}^{n \times r},
\qquad
U_*^\top U_* = I_r,
\qquad
\mathrm{span}(U_*) = T_*,
\]
and define the tangent and normal projectors by
\[
P_* := U_*U_*^\top,
\qquad
P_*^\perp := I - P_*.
\]
For any perturbation $h$, we therefore decompose
\[
h = h_{\parallel} + h_{\perp},
\qquad
h_{\parallel} := P_* h,
\qquad
h_{\perp} := P_*^\perp h.
\]

The core question is how the measurement operator acts on the tangent directions of this local manifold. If a tangent direction is weakly measured, two nearby prior-plausible images can remain nearly indistinguishable under $A$; if the tangent space is well observed, the measurements rule out such alternatives.

\begin{definition}[Local Measurement Quality]
For a tangent basis $U_*$ at $x_*$, define
\begin{equation}
\alpha(A;x_*)
:=
\sigma_{\min}(AU_*),
\label{eq:alpha-def}
\end{equation}
and, for $\tau > 0$,
\begin{equation}
\mathcal{S}_{\tau}(A;x_*)
:=
\log\det\!\bigl(U_*^\top A^\top A U_* + \tau I_r\bigr).
\label{eq:score-def}
\end{equation}
\end{definition}

The quantity $\alpha(A;x_*)$ is the conservative certificate used in the local stability bound: it measures the least observable tangent direction. The log-determinant score $\mathcal{S}_{\tau}(A;x_*)$ is its design-oriented companion: it rewards overall volume preservation on the local manifold and is better suited to greedy measurement selection. Thus $\alpha$ explains worst-case local ambiguity, while $\mathcal{S}_\tau$ operationalizes the same geometry for fixed and adaptive design.

\section{Local Stability from Measurement Observability}
\label{sec:theory}

This section answers the first question: for a fixed measurement operator $A$, how can its relationship with the local data manifold explain reconstruction feasibility? The claim is that $\alpha(A;x_*)$ controls the tangent component of reconstruction error. We state this as a local theorem under explicit assumptions. This is the appropriate level of abstraction for diffusion inverse solvers: the theorem identifies the measurement-manifold geometry that must be preserved, rather than proving global convergence for a particular sampler.

\begin{assumption}[Local Manifold Regularity]
\label{ass:local-manifold}
There exists a neighborhood $\mathcal{N}(x_*)$ in which the prior-relevant set is well approximated by a differentiable manifold with tangent basis $U_*$ and projectors $P_*,P_*^\perp$ defined in Section~\ref{sec:problem-setup}.
\end{assumption}

\begin{assumption}[Local Manifold-Aware Regularizer]
\label{ass:regularizer}
There exist constants $\mu_\perp > 0$, $L_\parallel \ge 0$ such that for every $x = x_* + h \in \mathcal{N}(x_*)$,
\begin{equation}
\mathcal{R}_{\theta}(x) - \mathcal{R}_{\theta}(x_*)
\ge
\frac{\mu_\perp}{2}\|h_\perp\|_2^2
\;-\;
\frac{L_\parallel}{2}\|h_\parallel\|_2^2.
\label{eq:regularizer-growth}
\end{equation}
\end{assumption}

Assumption~\ref{ass:regularizer} captures the local behavior expected from a generative prior: the regularizer penalizes motion away from the data manifold while allowing relatively flat motion along the manifold. Generator distance penalties, locally conservative score energies, and denoiser fixed-point penalties yield this tangent-normal form under standard local immersion, Hessian, or Jacobian conditions; details are in Appendix~\ref{app:proofs}.

\begin{assumption}[Approximate Local Optimality]
\label{ass:approx-min}
The reconstruction $\hat{x}$ returned by \eqref{eq:master-objective} lies in $\mathcal{N}(x_*)$ and satisfies
\begin{equation}
\Phi_{A,y,\lambda,\theta}(\hat{x})
\le
\Phi_{A,y,\lambda,\theta}(x_*) + \varepsilon
\label{eq:epsilon-opt}
\end{equation}
for some optimization error $\varepsilon \ge 0$.
\end{assumption}

\begin{theorem}[Local Stability under Tangent Observability]
\label{thm:main-stability}
Let Assumptions~\ref{ass:local-manifold}--\ref{ass:approx-min} hold. Define
\begin{equation}
\alpha_* := \alpha(A;x_*) = \sigma_{\min}(AU_*),
\qquad
\beta_* := \|A P_*^\perp\|_{2 \to 2},
\label{eq:alpha-beta}
\end{equation}
and let $\delta=\hat{x}-x_*=\delta_\parallel+\delta_\perp$. Then
\begin{equation}
\Bigl(\frac{\alpha_*^2}{8} - \frac{\lambda L_\parallel}{2}\Bigr)\|\delta_\parallel\|_2^2
\;+\;
\Bigl(\frac{\lambda \mu_\perp}{2} - \frac{\beta_*^2}{4}\Bigr)\|\delta_\perp\|_2^2
\;\le\;
\|e\|_2^2 + \varepsilon.
\label{eq:main-stability-bound}
\end{equation}
In particular, if $\alpha_*^2 > 4\lambda L_\parallel$ and $2\lambda\mu_\perp>\beta_*^2$, both tangent and normal errors are locally controlled.
\end{theorem}

\paragraph{Interpretation.}
The theorem separates two failure modes. The normal component $\delta_\perp$ is controlled by the prior's normal curvature $\mu_\perp$; if the regularizer does not reject off-manifold drift, the solver can leave the plausible set. The tangent component $\delta_\parallel$ is controlled by $\alpha_*=\sigma_{\min}(AU_*)$; if the measurements hide a locally plausible tangent direction, the prior cannot distinguish the true point from a nearby alternative on the manifold. This is the mathematical version of the hallucination concern: a plausible alternative is dangerous precisely when it is also nearly measurement-invisible.

\begin{corollary}[Pure Tangent Case]
\label{cor:pure-tangent}
If $\alpha(A;x_*)>0$, $\delta_\perp=0$, and $\mathcal{R}_\theta(\hat{x})\ge \mathcal{R}_\theta(x_*)$, then
\begin{equation}
\|\hat{x} - x_*\|_2
\le
\frac{2\sqrt{\|e\|_2^2 + \varepsilon}}{\alpha(A;x_*)}.
\label{eq:pure-tangent-bound}
\end{equation}
\end{corollary}

\begin{corollary}[Worst-Direction Design Criterion]
\label{cor:design}
Fix $\lambda$ and $L_\parallel$. For any two feasible measurement operators $A$ and $A'$ with
\[
\alpha(A';x_*)\ge \alpha(A;x_*),
\]
the tangent coefficient in \eqref{eq:main-stability-bound} for $A'$ is at least the tangent coefficient for $A$. Thus maximizing $\alpha(A;x_*)$ directly improves the worst observable tangent direction in the local bound.
\end{corollary}

The log-determinant score $\mathcal{S}_\tau(A;x)$ used in our algorithms is a practical surrogate for this worst-direction objective. It rewards volume preservation over the local tangent space and is more stable for greedy selection, but it should be interpreted as a design surrogate rather than as a theorem-level guarantee that the minimum singular value increases at every step.

The proof is a descent inequality for \eqref{eq:master-objective}: approximate optimality bounds $\|A\delta\|_2^2$ plus regularizer growth, and the decomposition $\delta=\delta_\parallel+\delta_\perp$ inserts $\alpha_*$ and $\beta_*$. Full details are provided in Appendix~\ref{app:proofs}.

\section{Posterior-Cloud Measurement Design}
\label{sec:posterior-cloud-design}

This section answers the second question: once a measurement-manifold compatibility score is available, can it be used to choose better measurements? The theory gives a conservative certificate through the smallest measured tangent direction, but practical acquisition design should also improve the overall coverage of the locally plausible set. We therefore use the theorem as motivation for the relevant geometry, and use a log-determinant objective as an operational surrogate for selecting measurements. The goal is not to claim that every greedy step maximizes the worst-case constant; it is to choose feasible measurements that preserve as much local posterior/manifold volume as possible under the acquisition budget.

For a fixed measurement family $\mathcal{A}_{\mathrm{feas}}$, the corresponding design objective is
\begin{equation}
\max_{A \in \mathcal{A}_{\mathrm{feas}}}
\;
\mathbb{E}_{x \sim p_{\mathrm{data}}}
\bigl[
\mathcal{S}_{\tau}(A;x)
\bigr],
\qquad
\mathcal{S}_{\tau}(A;x)
=
\log\det\!\bigl(U_x^\top A^\top A U_x + \tau I\bigr),
\label{eq:fixed-design-objective}
\end{equation}
with $\alpha(A;x)$ as the worst-direction counterpart. The determinant score is convenient for greedy design because it rewards volume preservation on the local tangent space and is less brittle than optimizing only the smallest singular value.

The main design variant is the sequential version of \eqref{eq:fixed-design-objective}. Given a total budget $m=m_1+m_2$, we first acquire an initial measurement operator $A_1$, reconstruct several coarse posterior samples conditioned on $y_1=A_1x_*+e_1$, estimate a local subspace $\widehat{U}$ from that posterior cloud, and greedily select additional feasible measurements $A_2$ by maximizing the incremental logdet score. The final reconstruction uses the combined operator
\begin{equation}
A_{\mathrm{final}}
=
\begin{bmatrix}
A_1\\
A_2
\end{bmatrix}.
\label{eq:combined-operator}
\end{equation}
The same template is used for row sampling, CT angle selection, and MR k-space line selection; only the feasible measurement set changes.

\paragraph{Sequential design procedure.}
Given a feasible measurement family and total budget, the posterior-cloud design proceeds in five steps: choose an initial structured operator $A_1$; run a short inverse solve conditioned on $y_1=A_1x_*+e_1$ to obtain coarse posterior samples; estimate a local subspace $\widehat{U}$ from this cloud; greedily add feasible measurements that maximize the incremental logdet score on $\widehat{U}$; and perform the final reconstruction using the completed operator $A_{\mathrm{final}}=[A_1;A_2]$.

Concretely, let $\mathcal{C}$ be the candidate set of remaining measurements and let $B(S)$ denote the operator formed by a selected set $S$. Starting from $S=\varnothing$, the second stage repeats
\begin{equation}
j^\star
\in
\arg\max_{j\in\mathcal{C}\setminus S}
\log\det\!\left(
\widehat{U}^{\top}
\begin{bmatrix}
A_1\\
B(S\cup\{j\})
\end{bmatrix}^{\top}
\begin{bmatrix}
A_1\\
B(S\cup\{j\})
\end{bmatrix}
\widehat{U}
+\tau I
\right),
\label{eq:greedy-logdet}
\end{equation}
then adds $j^\star$ to $S$. This is a greedy approximation, not a claim of global optimality. Its role is to make the theoretical quantity operational under scanner-like constraints.

The information budget is deliberately restricted. The method does not train a mask network, does not optimize a sampling policy over the training set, and does not use ground-truth test images. It uses the pretrained generative prior already required by the inverse solver and the posterior cloud produced from the current first-stage measurements. This distinction is central to the comparison with learned acquisition methods: those methods may be stronger in absolute performance, but they answer a different question because their sampling rule has seen training images.

\section{Experiments}
\label{sec:experiments}

The experiments ask whether changing the measurement operator alone can induce or remove hallucinations, whether the proposed geometry scores predict reconstruction behavior, and whether the same geometry can guide better measurements under a fixed budget. We study these questions only within structured measurement families rather than arbitrary dense matrices: row sampling chooses image rows, MNIST-CT chooses projection angles, and FastMRI\citep{zbontar2018fastmri} Cartesian sampling chooses phase-encoding lines in k-space. These settings expose the same measurement-geometry question under different acquisition constraints.

\paragraph{Solvers, baselines, and fairness.}
All comparisons fix the reconstruction solver within each setting and vary only the measurement rule. The MNIST row and CT experiments use a public pretrained MNIST DDPM with DPS-style reconstruction. The MR experiments use the Ambient Diffusion Posterior Sampling codebase \citep{aali2024ambient} with its pretrained diffusion model and the fastMRI data format. Detailed budgets, seeds, and solver settings are in Appendix~\ref{app:reproducibility}.

\subsection{MNIST Row Sampling}
\label{sec:mnist-row}

\begin{figure}[!t]
\centering
\includegraphics[width=0.98\linewidth]{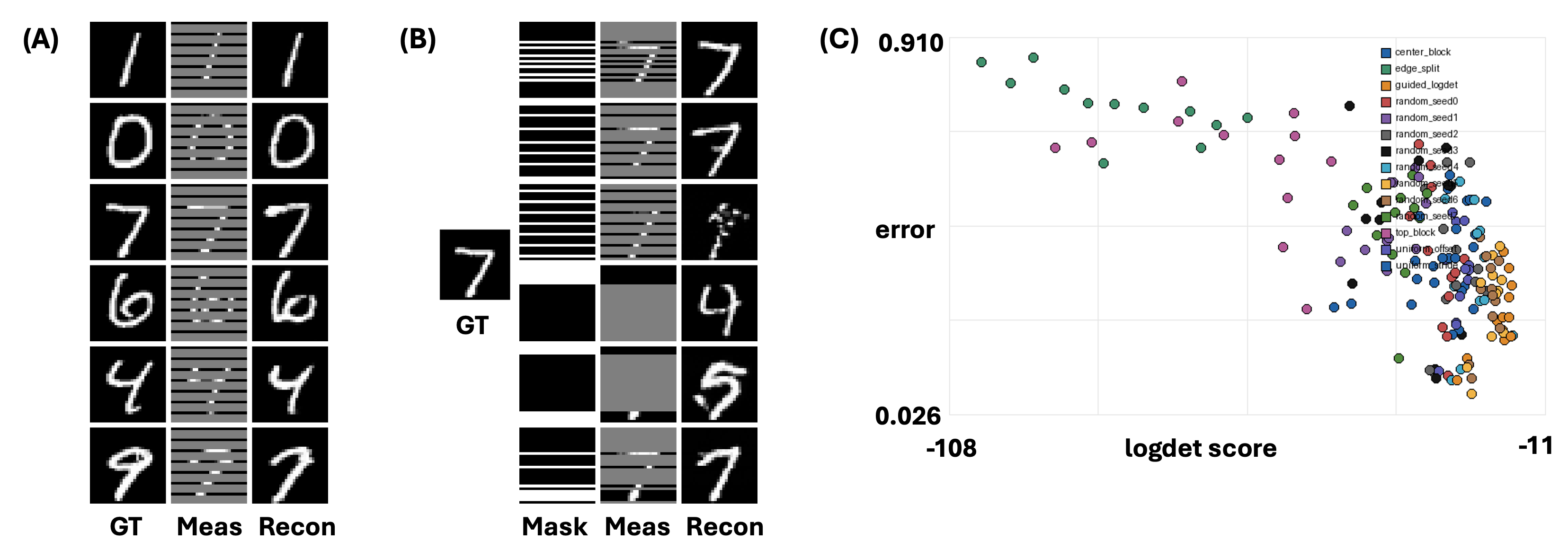}
\caption{MNIST row sampling: hallucination, fixed-mask design, and geometry-error relationship. (A) With a fixed stride-4 row operator, the same pretrained prior and DPS-style solver can reconstruct some digits correctly but can also complete a partially observed 9 as a plausible 7. Columns show ground truth, measured rows, and reconstruction. (B) For the same digit and row budget, different fixed masks produce different measurements and reconstructions, including severe hallucinations under poorly aligned masks. (C) Across fixed row operators and test images, higher logdet geometry scores tend to correspond to lower DPS relative error; the guided logdet mask lies in the favorable high-score, low-error region. Additional examples from the original galleries are provided in Appendix~\ref{app:additional-experiments}.}
\label{fig:mnist-row-fixed}
\label{fig:row-stage1}
\label{fig:row-fixed-design}
\end{figure}

The first experiment isolates the paper's central phenomenon. The prior, solver, and measurement budget are fixed; only the observed rows change. Figure~\ref{fig:mnist-row-fixed}(A) shows that a DPS-style solver can complete many partially observed digits, but can also reconstruct a label-9 image as a plausible 7. This is the failure mode motivating the paper: a visually reasonable output can be measurement-consistent because the operator fails to separate locally plausible alternatives.

We next ask whether the proposed geometry explains these differences before doing any adaptive design. For fixed seven-row operators, we estimate a local tangent basis by kNN-PCA and compute $\hat{\alpha}$ and logdet scores. Table~\ref{tab:geometry-correlation} shows that the geometry scores correlate with DPS error in the predicted direction: operators that preserve local tangent directions and volume tend to reconstruct better. The point is not exact error prediction from a finite PCA estimate, but evidence that measurement-manifold geometry is visible in solver behavior.

\begin{table}[!t]
\centering
\caption{Fixed MNIST row-operator geometry correlations. Each entry is the correlation with DPS relative error. Negative values support the theorem's directional prediction.}
\label{tab:geometry-correlation}
\begin{tabular}{lcc}
\toprule
Empirical geometry quantity & Pearson & Spearman \\
\midrule
$\hat{\alpha}_{\min}(A;x)$ vs. DPS relative error & -0.480 & -0.471 \\
$\hat{\alpha}_{\mathrm{mean}}(A;x)$ vs. DPS relative error & -0.605 & -0.551 \\
$\widehat{\mathcal{S}}_\tau(A;x)$ vs. DPS relative error & -0.584 & -0.558 \\
Observed tangent energy vs. DPS relative error & -0.566 & -0.544 \\
\bottomrule
\end{tabular}

\end{table}

\paragraph{Fixed row design.}
\label{sec:fixed-row-design}

The same score can be used constructively. Here the goal is not sample-specific adaptation yet, but a fixed acquisition rule that is better than generic row patterns. We greedily optimize a fixed seven-row operator on training anchors using the dataset-level logdet objective. Figure~\ref{fig:mnist-row-fixed}(B,C) shows both sides of the evidence: mask choice changes the failure modes of the same solver, and the guided design lies in the favorable high-score, low-error region of the logdet-error scatter.

\paragraph{Adaptive row design.}
\label{sec:adaptive-row-design}

The final MNIST row experiment tests the paper's main design claim: after an initial acquisition, the measurement operator can be customized to the current image without using its ground truth. The first stage acquires the guided seven-row prefix. The second stage adds four rows by greedy logdet selection using either kNN-PCA around the coarse reconstruction or a posterior cloud generated from the current measurements. All final methods use the same total number of rows.

\begin{table}[!t]
\centering
\caption{MNIST row adaptive acquisition. Each method uses 11 total observed rows.}
\label{tab:row-main}
\resizebox{\linewidth}{!}{\begin{tabular}{lcccccc}
\toprule
Metric & Adaptive posterior cloud & Adaptive kNN-PCA & Guided fixed & Uniform offset & Uniform stride & Best random \\
\midrule
Relative error $\downarrow$ & $\mathbf{0.1501}$ & $0.1718$ & $0.1827$ & $0.2029$ & $0.2496$ & $0.2169$ \\
PSNR $\uparrow$ & $\mathbf{23.33}$ & $22.28$ & $21.63$ & $20.70$ & $19.09$ & $20.17$ \\
\bottomrule
\end{tabular}
}
\end{table}

\begin{figure}[!t]
\centering
\includegraphics[width=0.8\linewidth]{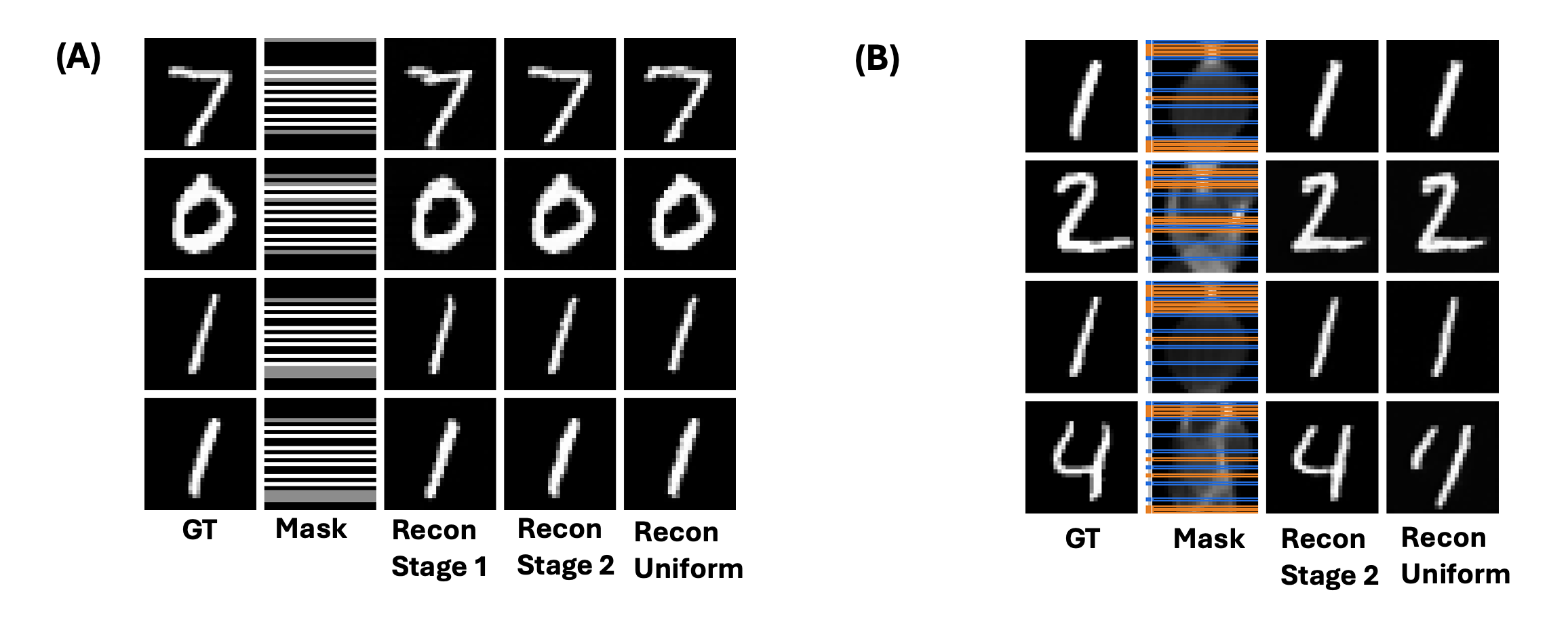}
\caption{Adaptive measurement design in row sampling and CT. (A) MNIST row posterior-cloud adaptation. Columns show ground truth, the final row mask, the stage-1 reconstruction from the initial rows, the stage-2 reconstruction after adding posterior-cloud-selected rows, and a uniform-row baseline. The selected second-stage rows vary across images, reflecting sample-specific posterior geometry. (B) MNIST-CT posterior-cloud angle selection for the $7+7$ setting. The mask is displayed in CT measurement space, i.e., sinogram space: each horizontal row corresponds to the detector measurements collected at one projection angle. Blue rows mark the initial stage-1 angles, and orange rows mark the angles added in stage 2 from the posterior cloud. The last two columns compare adaptive stage-2 reconstruction with a uniform-angle baseline.}
\label{fig:adaptive-row-ct}
\label{fig:row-adaptive-gallery}
\label{fig:ct-crop}
\end{figure}

Table~\ref{tab:row-main} reports the adaptive row results. Posterior-cloud adaptation performs best among the tested row rules, and Figure~\ref{fig:adaptive-row-ct}(A) shows why this is not simply a fixed mask in disguise: the second-stage rows vary with the sample. This is the experimental counterpart of the theory: the relevant geometry is local to the current posterior, so useful additional measurements need not be the same for every image.

\subsection{MNIST-CT}
\label{sec:mnist-ct}

MNIST-CT replaces row measurements with a dense parallel-beam Radon operator. We treat it as a structured transfer check rather than a full CT protocol-design study: real CT acquisition couples view selection with dose allocation, detector geometry, and clinical constraints, while here each candidate angle has equal dose and the design chooses a fixed number of angles. We compare five angle rules: \emph{Adaptive} adds second-stage angles from a posterior cloud; \emph{Guided fixed} uses the same logdet principle but fixes one angle set before seeing each test image; \emph{Uniform} and \emph{Uniform offset} use evenly spaced grids; and \emph{Random mean} averages over random angle sets. Table~\ref{tab:ct-main} shows that adaptive angle design ranks first in all reported settings. Figure~\ref{fig:adaptive-row-ct}(B) visualizes the same policy in CT measurement space, and Appendix~\ref{app:additional-experiments} gives the full gallery.

\begin{table}[!t]
\centering
\caption{MNIST-CT posterior-cloud angle selection on $n=100$ test images. Entries are mean DPS relative errors; lower is better. Each setting uses a fixed number of initial angles and then adds the same number of second-stage angles.}
\label{tab:ct-main}
\resizebox{\linewidth}{!}{\begin{tabular}{lccccc}
\toprule
Setting & Adaptive & Guided fixed & Uniform & Uniform offset & Random mean \\
\midrule
CT $6+6$ angles & 0.3360 & 0.3393 & 0.3482 & 0.3508 & 0.3729 \\
CT $7+7$ angles & 0.3145 & 0.3212 & 0.3231 & 0.3259 & 0.3486 \\
CT $8+8$ angles & 0.2979 & 0.3061 & 0.3105 & 0.3111 & 0.3246 \\
\bottomrule
\end{tabular}
}
\end{table}

\subsection{FastMRI Cartesian Sampling}
\label{sec:fastmri}

We use Cartesian line sampling in the FastMRI image domain at acceleration $R=6$ with 53 phase-encoding lines and a 20-line ACS region. The non-learned baselines preserve ACS and vary only the outer k-space selection, following calibration-aware and variable-density MRI sampling principles \citep{lustig2007sparse,zbontar2018fastmri,dwork2021fast,cheng2019compressed}: ACS + equispaced, ACS + random, ACS + Gaussian VDS, and ACS + VD-Poisson-like. Our two adaptive variants use the same final line budget but split acquisition into an equispaced or VDS/Poisson-like prefix followed by posterior-cloud-selected lines. We do not include learned mask policies as main baselines because training-set adaptive sampling and jointly learned acquisition methods \citep{ravishankar2011adaptive,gozcu2018learning,bahadir2020loupe} optimize masks or predictors over a reconstruction training distribution, whereas our comparison isolates test-time measurement design without training a mask policy.

\begin{table}[!t]
\centering
\caption{MR Cartesian sampling at $R=6$ over 100 fastMRI-domain cases. All methods use 53 phase-encoding lines and 20 ACS lines where applicable.}
\label{tab:mr}
\resizebox{\linewidth}{!}{\begin{tabular}{lccc}
\toprule
Method & NRMSE $\downarrow$ & SSIM $\uparrow$ & PSNR $\uparrow$ \\
\midrule
ACS + equispaced & $0.1187 \pm 0.0201$ & $0.9055 \pm 0.0278$ & $33.18 \pm 1.65$ \\
ACS + random & $0.1124 \pm 0.0197$ & $0.9107 \pm 0.0262$ & $33.74 \pm 1.70$ \\
Ours, equispaced prefix + adaptive & $0.0948 \pm 0.0164$ & $0.9269 \pm 0.0219$ & $35.14 \pm 1.68$ \\
ACS + Gaussian VDS & $0.0781 \pm 0.0123$ & $0.9426 \pm 0.0172$ & $36.93 \pm 1.60$ \\
ACS + VD-Poisson-like & $0.0662 \pm 0.0103$ & $0.9528 \pm 0.0153$ & $38.24 \pm 1.60$ \\
Ours, VDS prefix + adaptive & $\mathbf{0.0620 \pm 0.0092}$ & $\mathbf{0.9574 \pm 0.0143}$ & $\mathbf{38.80 \pm 1.55}$ \\
\bottomrule
\end{tabular}
}
\end{table}

\begin{figure}[!t]
\centering
\includegraphics[width=0.8\linewidth]{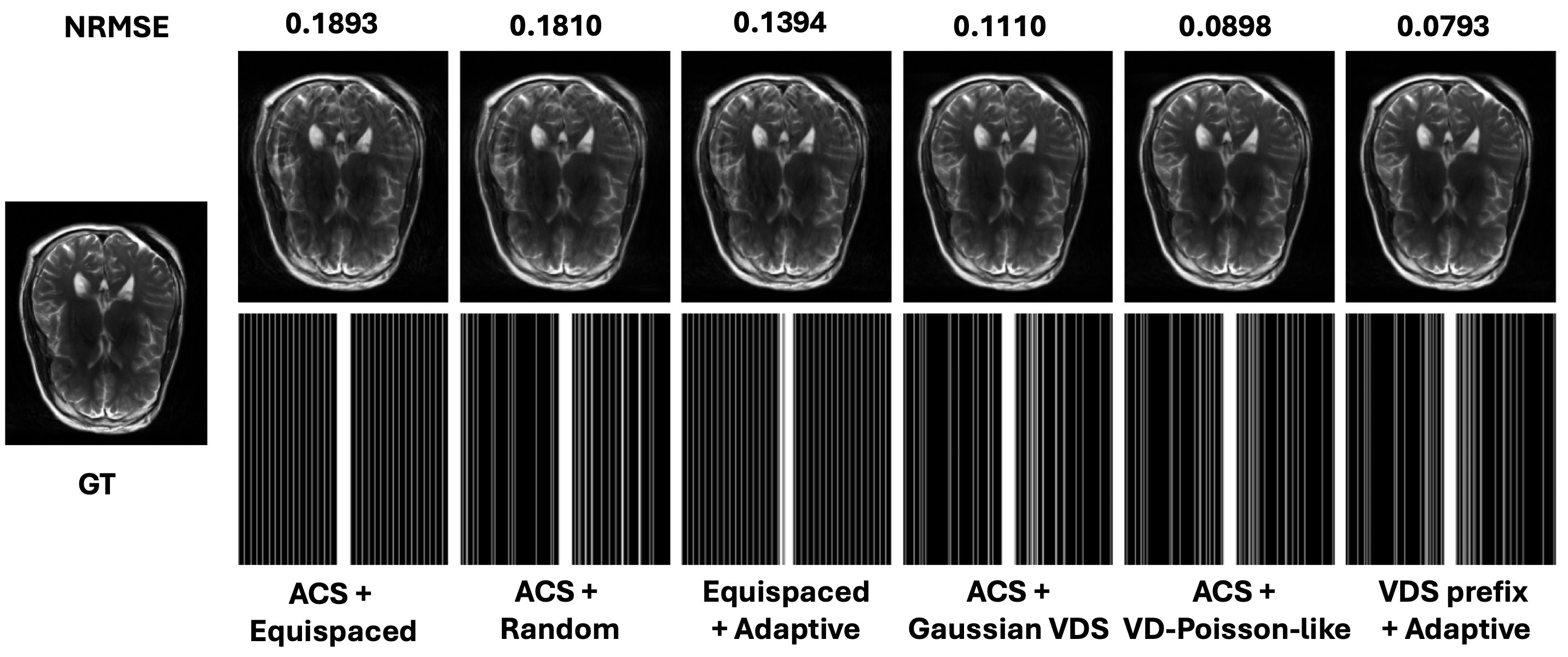}
\caption{Representative MR Cartesian sampling example. The left panel shows the ground truth; the remaining panels compare reconstructions, masks, and NRMSE for six sampling rules under the same $R=6$ line budget. The masks show selected phase-encoding lines in Cartesian k-space. Because the experiment is set at a realistic acceleration with ACS-preserving masks and a strong posterior solver, the reconstructions are not expected to show dramatic anatomical differences; the relevant comparison is the residual artifact level, quantitative error, and how the sampling pattern changes across equispaced, random, variable-density, Poisson-like, and adaptive designs. Additional MR examples are in Appendix~\ref{app:additional-experiments}.}
\label{fig:mr-sample}
\end{figure}

Table~\ref{tab:mr} reports the MR metrics. The strongest comparator is the ACS-preserving VD-Poisson-like baseline, so the margin is expectedly smaller than in settings where the candidate operators differ more visibly. The relevant claim is therefore modest but important: posterior-cloud measurement design can improve a realistic Cartesian operator even when compared against well-tuned non-learned ACS-preserving baselines.

\section{Discussion, Limitations, and Conclusion}

\textbf{Discussion.}
This paper develops a compressed-sensing-style view of generative inverse problems: once the prior and solver are fixed, the measurement operator remains a first-class determinant of reconstruction reliability. Hallucination can be induced by an operator that leaves plausible manifold directions weakly observed, and the tangent observability constant $\alpha(A;x)=\sigma_{\min}(AU_x)$ captures this local measurement-prior interaction in a stability bound. The experiments support the resulting geometry-to-design chain across row sampling, MNIST-CT angle selection, and FastMRI Cartesian sampling: changing structured measurements changes failure modes, geometry scores track reconstruction behavior, and posterior-cloud design improves fixed-budget acquisition without training a mask policy.

\textbf{Limitations.}
The claims are intentionally local and directional rather than global. The index is a practical volume surrogate rather than an exact error predictor or a guarantee that every greedy step improves the worst singular direction. The comparison also reflects a different application setting from fully learned acquisition policies: many learned methods leverage large training datasets and joint acquisition–reconstruction training, whereas our framework focuses on adaptive test-time measurement design from posterior geometry. Finally, adaptive acquisition introduces additional computation through a first-stage posterior-geometry solve, and practical deployment may require tighter integration with hardware and acquisition constraints.

\textbf{Conclusion.}
The main conclusion is that trustworthiness in generative inverse problems is not only a property of the prior or sampler; it also depends on whether the measurement operator resolves the locally plausible alternatives induced by that prior. This viewpoint gives both an explanatory index and a design principle for structured acquisition. Reducing the overhead of posterior-cloud estimation and extending the framework to richer clinical acquisition constraints are natural next steps.

\section*{Acknowledgment}

This work was supported by the National Institutes of Health (NIH) under award number R01HL159183 and the National Science Foundation (NSF) AI Institute under award number 2112085.

\bibliographystyle{plainnat}
\bibliography{references}

\clearpage
\appendix
\section{Proof Details}
\label{app:proofs}

This appendix collects the proof details that are abbreviated in Section~\ref{sec:theory}.

\subsection{Regularizer Models}

For the generator-induced penalty, assume that $G_\theta$ is a local immersion at $z_*$ and that $x_*=G_\theta(z_*)$. In a sufficiently small neighborhood, the image of $G_\theta$ is a $C^2$ embedded manifold. The minimizing latent variable in
\[
\mathcal{R}_{\theta}^{\mathrm{gen}}(x)
=
\min_z \frac{1}{2\sigma_g^2}\|x-G_\theta(z)\|_2^2+\psi(z)
\]
is the local projection of $x$ onto the range of $G_\theta$, up to the locally flat latent penalty $\psi$. The squared distance to a $C^2$ manifold has the expansion
\[
\mathrm{dist}(x_*+h,\mathcal{M})^2
=
\|h_\perp\|_2^2 + O(\|h\|_2^3).
\]
This yields
\[
\mathcal{R}_{\theta}^{\mathrm{gen}}(x_*+h)-\mathcal{R}_{\theta}^{\mathrm{gen}}(x_*)
=
\frac{1}{2\sigma_g^2}\|h_\perp\|_2^2+O(\|h\|_2^3),
\]
which implies Assumption~\ref{ass:regularizer} in a sufficiently small neighborhood. Indeed, if $|O(\|h\|_2^3)|\le C\|h\|_2^3$ and $\|h\|_2\le \rho$, then the cubic remainder is bounded below by $-C\rho(\|h_\parallel\|_2^2+\|h_\perp\|_2^2)$. Choosing $\rho$ small enough preserves a positive normal coefficient and leaves a finite tangent slack.

For the score-induced penalty, suppose $\mathcal{R}_{\theta}^{\mathrm{score}}=-\varphi_\theta$ is $C^2$ near $x_*$ and $\nabla\mathcal{R}_{\theta}^{\mathrm{score}}(x_*)=0$. Taylor expansion gives
\[
\mathcal{R}_{\theta}^{\mathrm{score}}(x_*+h)-\mathcal{R}_{\theta}^{\mathrm{score}}(x_*)
=
\frac{1}{2}h^\top \nabla^2\mathcal{R}_{\theta}^{\mathrm{score}}(x_*)h+o(\|h\|_2^2).
\]
Let $H=\nabla^2\mathcal{R}_{\theta}^{\mathrm{score}}(x_*)$. If the Hessian is positive in normal directions, not too negative in tangent directions, and has controlled tangent-normal coupling, namely
\[
P_*^\perp H P_*^\perp \succeq \mu_\perp I,
\qquad
P_* H P_* \succeq -L_0 I,
\qquad
\|P_* H P_*^\perp\|_{2\to 2}\le \gamma,
\]
then Young's inequality gives, for any $\eta>0$,
\[
h^\top Hh
\ge
\bigl(\mu_\perp-\eta\gamma\bigr)\|h_\perp\|_2^2
-
\bigl(L_0+\gamma/\eta\bigr)\|h_\parallel\|_2^2.
\]
After shrinking the neighborhood to absorb the $o(\|h\|_2^2)$ term, the desired tangent-normal lower bound follows whenever $\eta$ is chosen so that $\mu_\perp-\eta\gamma>0$.

For the denoiser-induced penalty, write $J=J_{D_\theta}(x_*)$. If $D_\theta(x_*)=x_*$, then
\[
x_*+h-D_\theta(x_*+h)=(I-J)h+o(\|h\|_2).
\]
Thus
\[
\mathcal{R}_{\theta}^{\mathrm{denoise}}(x_*+h)
=
\frac{1}{2\eta^2}\|(I-J)h\|_2^2+o(\|h\|_2^2).
\]
If there exist constants $c_\perp>0$ and $c_\parallel\ge 0$ such that
\[
\|(I-J)h\|_2^2
\ge
c_\perp^2\|h_\perp\|_2^2
-
c_\parallel^2\|h_\parallel\|_2^2
\]
for all sufficiently small $h$, then the lower bound holds with $\mu_\perp$ proportional to $c_\perp^2/\eta^2$ and $L_\parallel$ proportional to $c_\parallel^2/\eta^2$, after absorbing the $o(\|h\|_2^2)$ term locally. These three examples justify the tangent-normal growth model used in Assumption~\ref{ass:regularizer}; they do not claim that every trained diffusion model globally satisfies the assumption.

\subsection{Descent Inequality and Stability Bound}

Let $\delta=\hat{x}-x_*$. Approximate local optimality gives
\[
\frac{1}{2}\|A\hat{x}-y\|_2^2+\lambda\mathcal{R}_\theta(\hat{x})
\le
\frac{1}{2}\|Ax_*-y\|_2^2+\lambda\mathcal{R}_\theta(x_*)+\varepsilon.
\]
Using $y=Ax_*+e$ and $\hat{x}=x_*+\delta$,
\[
\frac{1}{2}\|A\delta-e\|_2^2-\frac{1}{2}\|e\|_2^2
+\lambda\left(\mathcal{R}_\theta(x_*+\delta)-\mathcal{R}_\theta(x_*)\right)
\le \varepsilon.
\]
The inequality
\[
\frac{1}{2}\|A\delta-e\|_2^2-\frac{1}{2}\|e\|_2^2
\ge
\frac{1}{4}\|A\delta\|_2^2-\|e\|_2^2
\]
implies the basic descent inequality
\begin{equation}
\frac{1}{4}\|A\delta\|_2^2
\;+\;
\lambda\bigl(\mathcal{R}_{\theta}(x_*+\delta)-\mathcal{R}_{\theta}(x_*)\bigr)
\le
\|e\|_2^2 + \varepsilon.
\label{eq:appendix-descent}
\end{equation}

Decompose $\delta=\delta_\parallel+\delta_\perp$. The elementary inequality
\[
\|a+b\|_2^2 \ge \frac{1}{2}\|a\|_2^2-\|b\|_2^2
\]
with $a=A\delta_\parallel$ and $b=A\delta_\perp$ gives
\[
\|A\delta\|_2^2
\ge
\frac{1}{2}\|A\delta_\parallel\|_2^2-\|A\delta_\perp\|_2^2.
\]
By definition,
\[
\|A\delta_\parallel\|_2\ge \alpha_*\|\delta_\parallel\|_2,
\qquad
\|A\delta_\perp\|_2\le \beta_*\|\delta_\perp\|_2.
\]
Substituting these two bounds and Assumption~\ref{ass:regularizer} into \eqref{eq:appendix-descent} yields
\[
\left(\frac{\alpha_*^2}{8}-\frac{\lambda L_\parallel}{2}\right)\|\delta_\parallel\|_2^2
+
\left(\frac{\lambda\mu_\perp}{2}-\frac{\beta_*^2}{4}\right)\|\delta_\perp\|_2^2
\le
\|e\|_2^2+\varepsilon.
\]
This is Theorem~\ref{thm:main-stability}. For Corollary~\ref{cor:pure-tangent}, use \eqref{eq:appendix-descent} directly: if $\alpha_*>0$, $\delta_\perp=0$, and $\mathcal{R}_\theta(\hat{x})\ge \mathcal{R}_\theta(x_*)$, then
\[
\frac{1}{4}\|A\delta\|_2^2
\le
\|e\|_2^2+\varepsilon.
\]
Since $\delta=\delta_\parallel$, $\|A\delta\|_2\ge \alpha_*\|\delta\|_2$, giving
\[
\|\delta\|_2
\le
\frac{2\sqrt{\|e\|_2^2+\varepsilon}}{\alpha_*}.
\]
Corollary~\ref{cor:design} follows because the tangent-side coefficient in \eqref{eq:main-stability-bound} is monotone in $\alpha_*^2$ when $\lambda$ and $L_\parallel$ are fixed. The logdet score used later is a volume-preserving surrogate for this worst-direction criterion, not a guarantee that $\alpha_*$ increases at every greedy step.

\section{Experimental Setting and Additional Results}
\label{app:reproducibility}
\label{app:additional-experiments}

This section records implementation details and additional visual evidence for the experiments. The central fairness rule is unchanged across settings: within each experiment family, all methods use the same reconstruction solver and total measurement budget, and only the structured measurement rule changes. Experiments were run on NVIDIA A100 GPUs.

\paragraph{Controlled MNIST settings.}
The MNIST row and CT experiments use the public pretrained MNIST DDPM \texttt{1aurent/ddpm-mnist}. Data are loaded from the MNIST cache under \texttt{Experiment/datasets/mnist\_cache}; row experiments use balanced test subsets, and the main adaptive row result uses 100 test images with 10 per digit. Row reconstruction uses a DPS-style solver with hard data consistency and guidance scale $0.12$. The fixed row geometry study uses seven observed rows, 13 fixed row operators, 12 balanced test images, kNN-PCA tangent estimation with 128 local neighbors, tangent dimension 16, and 100 inference steps. The fixed row design greedily optimizes a dataset-level logdet objective over seven rows using 64 training anchors, 128 local neighbors, tangent dimension 16, and ridge $\tau=10^{-3}$. Adaptive row design starts from seven guided rows and adds four rows; the posterior-cloud variant uses four posterior samples, 75 coarse/cloud inference steps, and 100 final inference steps. MNIST-CT uses parallel-beam Radon measurements with 28 candidate angles and 28 detector bins. The main CT table reports posterior-cloud angle selection on 100 balanced test images, tangent dimension 12, ridge $\tau=10^{-3}$, posterior cloud size 4, and 50 inference steps for cloud, coarse, and final reconstructions; each candidate angle is treated as having equal dose.

\paragraph{FastMRI Cartesian setting.}
The MR experiment uses the Ambient Diffusion Posterior Sampling implementation \citep{aali2024ambient}, specifically \texttt{solve\_inverse\_adps.py} with \texttt{--method edm}, \texttt{--l\_type DPS}, \texttt{--l\_ss 1}, and \texttt{--S\_churn 0}, together with the pretrained \texttt{models/edm/supervised\_R=1} network. The model is trained in the fastMRI domain, while evaluation samples are generated unconditionally rather than selected from the training set. The forward model is Cartesian k-space line sampling on $384\times320$ images at acceleration $R=6$ with 53 phase-encoding lines; all ACS-preserving methods use 20 central ACS lines. We evaluate 100 MR-domain samples, \texttt{sample\_\{0..99\}.pt}, using NRMSE, SSIM, and PSNR. Single-stage baselines are ACS + equispaced, ACS + random, ACS + Gaussian VDS, and ACS + VD-Poisson-like masks. Gaussian and VD-Poisson-like masks use VDS scale $0.25$, and the Poisson-like variant uses minimum gap 2; randomized masks use seeds $\{0,1,2\}$ and are averaged per sample. Final reconstructions use 300 DPS steps with reconstruction seed 15 and latent seed 15. Adaptive policies use a shorter first-stage posterior-cloud solve with 180 DPS steps and latent seeds $\{10,\ldots,17\}$, followed by the same 300-step final solve. The equispaced-prefix adaptive policy starts from 36 lines, while the VDS-prefix adaptive policy starts from 37 lines and adds 16 adaptive lines to keep the low-/high-frequency balance close to the corresponding strong baseline. No method uses ground-truth test images for mask design, and the proposed policy does not train a mask network or optimize a sampling policy over the reconstruction training set.

\subsection{MNIST Row Examples}

\begin{figure}[h]
\centering
\includegraphics[width=0.82\linewidth,height=0.62\textheight,keepaspectratio]{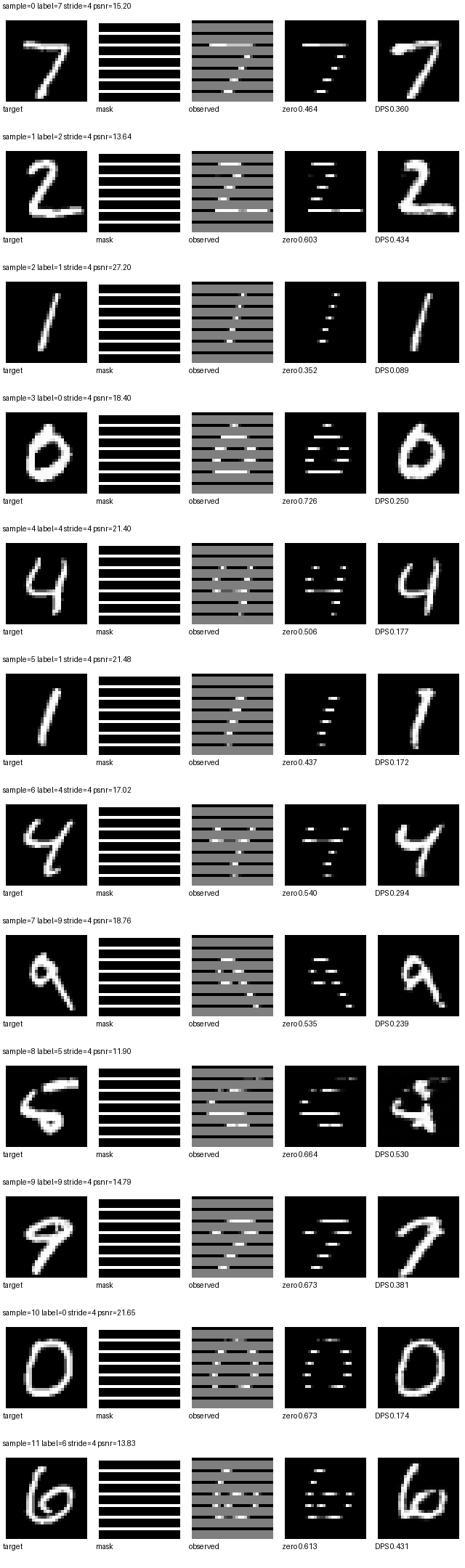}
\caption{Additional MNIST row-inpainting examples under the fixed stride-4 row operator used in Figure~\ref{fig:row-stage1}. The gallery shows that the same pretrained prior and DPS-style solver can produce both accurate reconstructions and plausible but incorrect completions when the observed rows do not distinguish nearby digit alternatives.}
\label{fig:appendix-row-stage1-gallery-copy}
\end{figure}

\begin{figure}[h]
\centering
\includegraphics[width=0.82\linewidth,height=0.62\textheight,keepaspectratio]{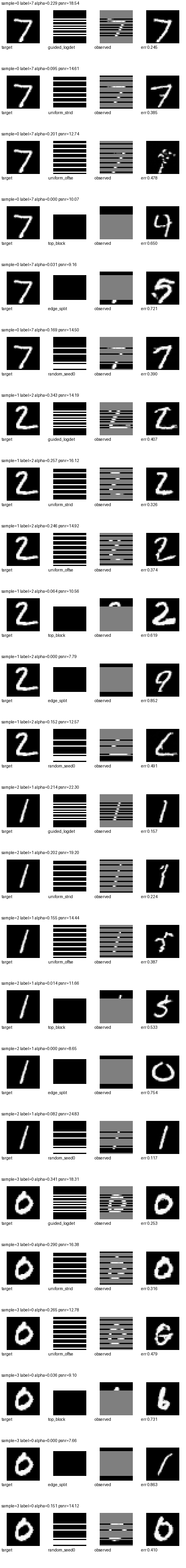}
\caption{Additional fixed-row design examples corresponding to Figure~\ref{fig:row-fixed-design}(a). Each block compares reconstruction behavior under representative row masks with the same measurement budget, illustrating how mask choice changes the failure modes of the same generative inverse solver.}
\label{fig:appendix-fixed-row-gallery-copy}
\end{figure}

\begin{figure}[h]
\centering
\includegraphics[width=0.82\linewidth]{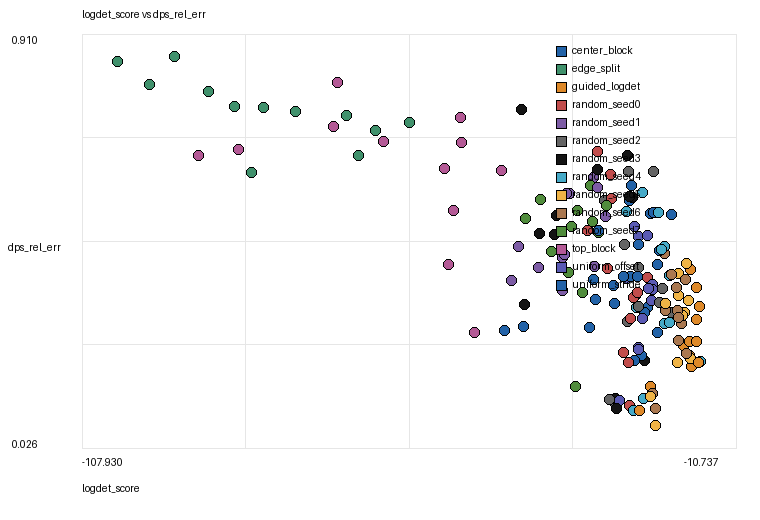}
\caption{Logdet score versus DPS relative error for the fixed-row design study, corresponding to Figure~\ref{fig:row-fixed-design}(b). The guided logdet mask lies in the high-score, low-error region, while poor structured masks tend to have lower geometry scores and larger reconstruction error.}
\label{fig:appendix-fixed-row-scatter-copy}
\end{figure}

\begin{figure}[h]
\centering
\includegraphics[width=0.75\linewidth]{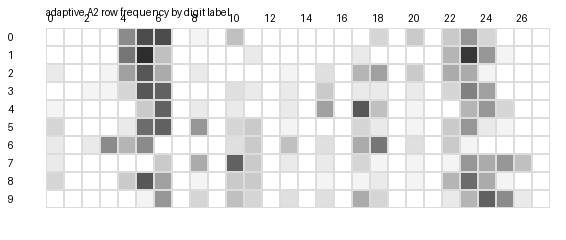}
\caption{Second-stage row-selection heatmap for posterior-cloud adaptive MNIST acquisition. The selected rows are concentrated around informative digit strokes but remain sample dependent.}
\label{fig:appendix-row-heatmap}
\end{figure}

\begin{figure}[h]
\centering
\includegraphics[width=0.92\linewidth,height=0.72\textheight,keepaspectratio]{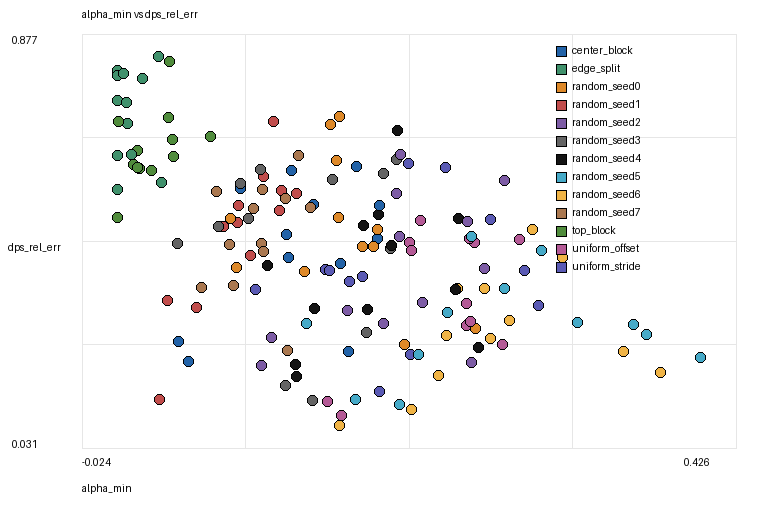}
\caption{Additional fixed-row geometry diagnostic: empirical $\hat{\alpha}_{\min}$ versus DPS relative error.}
\label{fig:appendix-alpha-scatter}
\end{figure}

This subsection provides additional MNIST row-sampling visualizations supporting Figure~\ref{fig:mnist-row-fixed} and Figure~\ref{fig:adaptive-row-ct}(A). Figure~\ref{fig:appendix-row-stage1-gallery-copy} shows additional row-inpainting outcomes under a fixed stride operator, while Figures~\ref{fig:appendix-fixed-row-gallery-copy} and~\ref{fig:appendix-fixed-row-scatter-copy} show the fixed-mask design gallery and the corresponding logdet-error diagnostic. Figures~\ref{fig:appendix-row-heatmap} and~\ref{fig:appendix-alpha-scatter} provide additional adaptive-selection and geometry-score diagnostics.

\subsection{MNIST-CT Results}

\begin{figure}[h]
\centering
\includegraphics[width=0.92\linewidth,height=0.72\textheight,keepaspectratio]{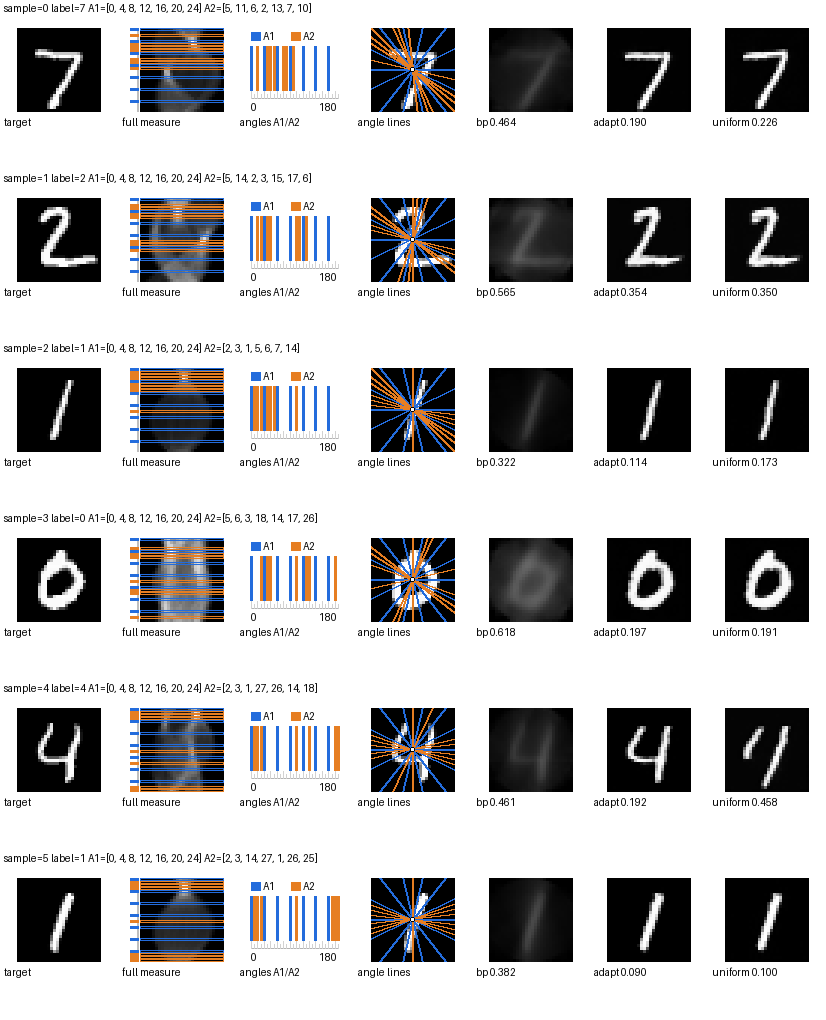}
\caption{MNIST-CT posterior-cloud angle-selection gallery for the $7+7$ run.}
\label{fig:appendix-ct-gallery}
\end{figure}

This subsection records the expanded CT angle-selection results. Table~\ref{tab:appendix-ct} includes the larger $7+7$ diagnostic run in addition to the main $n=100$ settings, and Figure~\ref{fig:appendix-ct-gallery} shows the corresponding $7+7$ posterior-cloud angle-selection gallery. The separate cropped CT figure from earlier drafts is omitted because its examples are now redundant with Figure~\ref{fig:adaptive-row-ct}(B).

\begin{table}[h]
\centering
\caption{Large MNIST-CT posterior-cloud angle-selection results. Entries are mean DPS relative errors except the win-rate column.}
\label{tab:appendix-ct}
\resizebox{\linewidth}{!}{\begin{tabular}{lcccccc}
\toprule
Setting & Adaptive & Guided fixed & Uniform & Uniform offset & Random mean & Win vs. guided \\
\midrule
CT $6+6$, posterior cloud, $n=100$ & 0.3360 & 0.3393 & 0.3482 & 0.3508 & 0.3729 & 0.56 \\
CT $7+7$, posterior cloud, $n=100$ & 0.3145 & 0.3212 & 0.3231 & 0.3259 & 0.3486 & 0.54 \\
CT $8+8$, posterior cloud, $n=100$ & 0.2979 & 0.3061 & 0.3105 & 0.3111 & 0.3246 & 0.57 \\
CT $7+7$, posterior cloud, $n=200$ & 0.3117 & 0.3262 & 0.3176 & 0.3212 & 0.3439 & 0.65 \\
\bottomrule
\end{tabular}
}
\end{table}

\subsection{FastMRI Cartesian Examples}

\begin{figure}[h]
\centering
\includegraphics[width=0.92\linewidth,height=0.72\textheight,keepaspectratio]{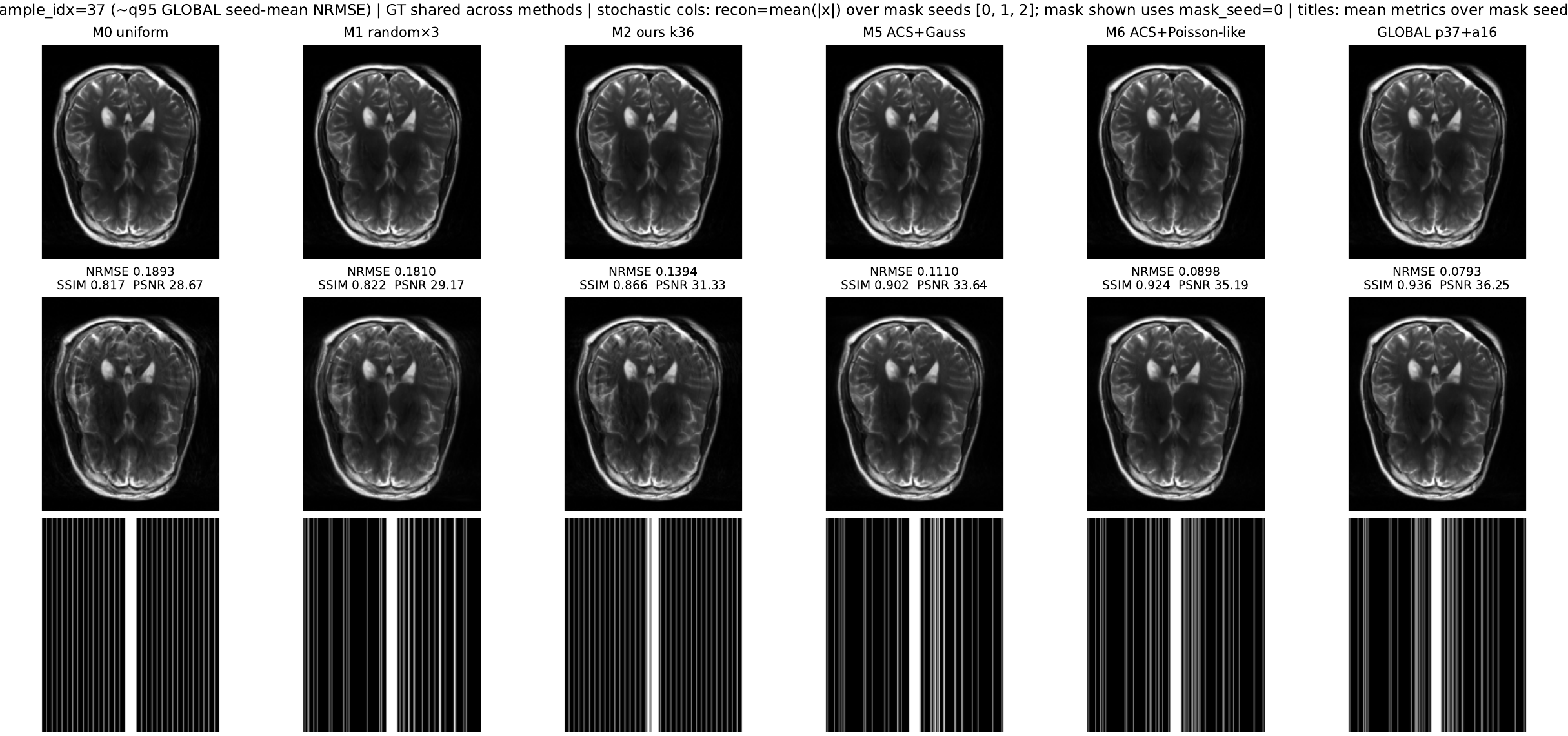}
\caption{Additional copy of the representative MR visualization in Figure~\ref{fig:mr-sample}, sample 037. The panels compare ground truth, reconstructions, and masks across six Cartesian sampling methods under the same total line budget.}
\label{fig:appendix-mr-037-copy}
\end{figure}

\begin{figure}[h]
\centering
\includegraphics[width=0.92\linewidth,height=0.72\textheight,keepaspectratio]{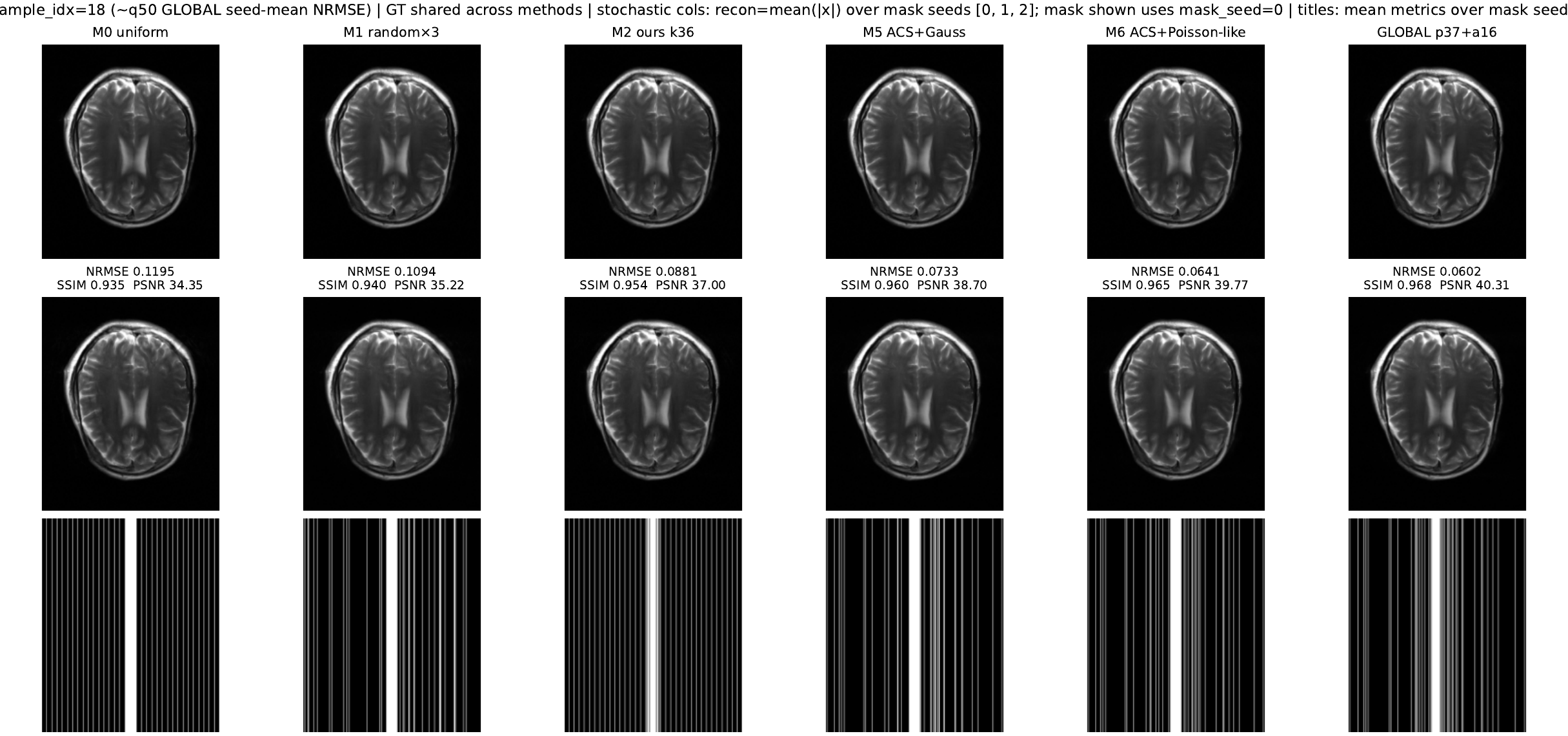}
\caption{Additional MR example, sample 018.}
\label{fig:appendix-mr-018}
\end{figure}

\begin{figure}[h]
\centering
\includegraphics[width=0.92\linewidth,height=0.72\textheight,keepaspectratio]{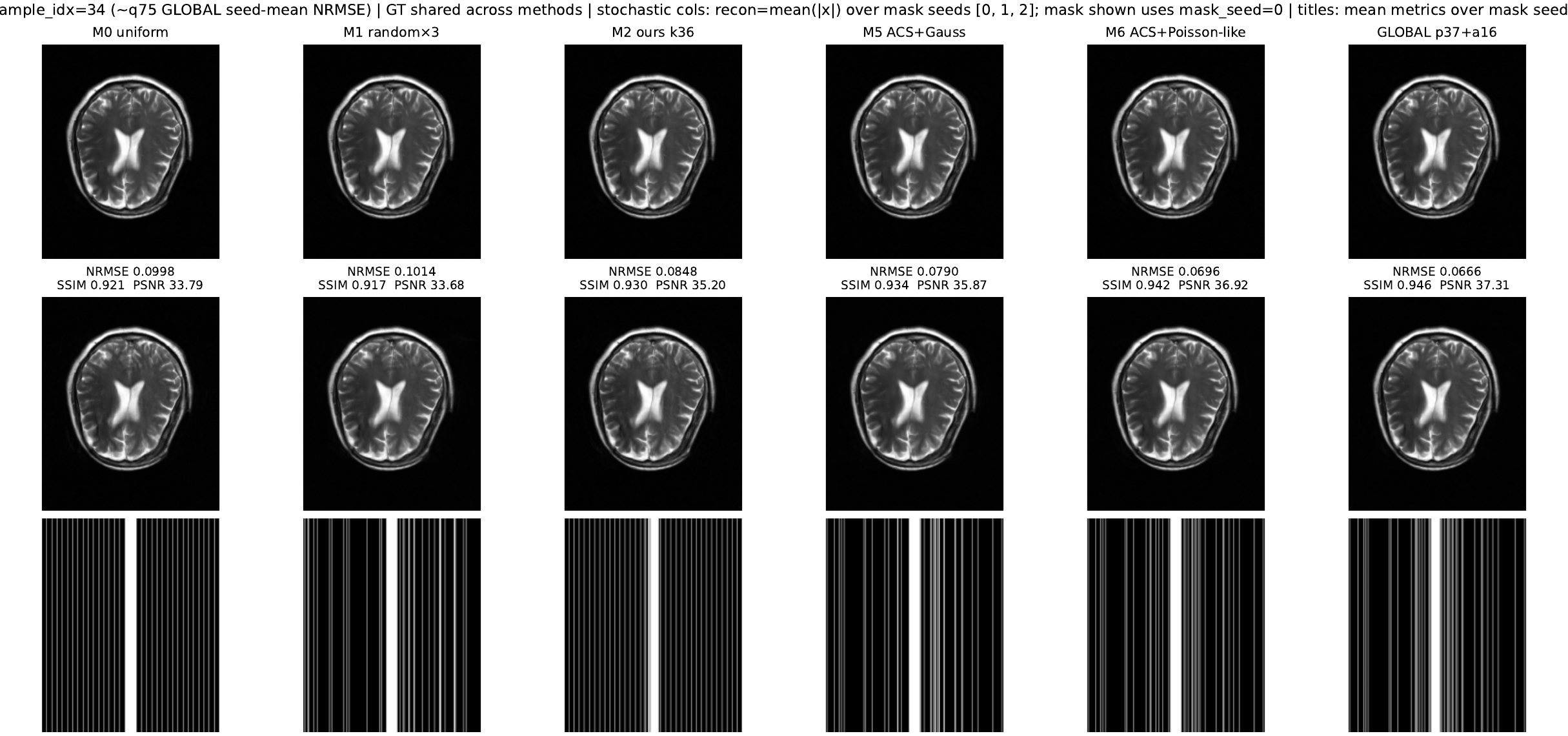}
\caption{Additional MR example, sample 034.}
\label{fig:appendix-mr-034}
\end{figure}

\begin{figure}[h]
\centering
\includegraphics[width=0.92\linewidth,height=0.72\textheight,keepaspectratio]{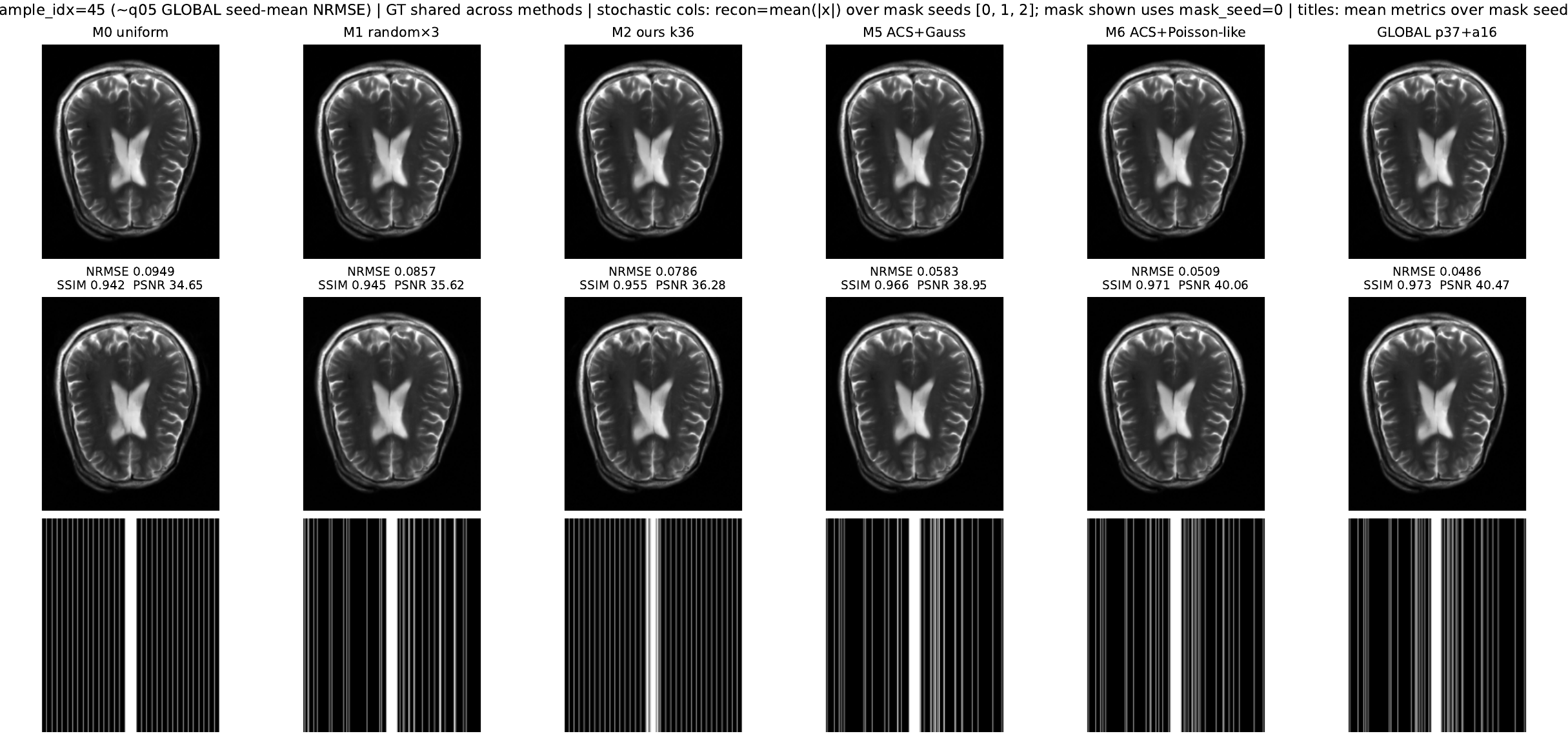}
\caption{Additional MR example, sample 045.}
\label{fig:appendix-mr-045}
\end{figure}

\begin{figure}[h]
\centering
\includegraphics[width=0.92\linewidth,height=0.72\textheight,keepaspectratio]{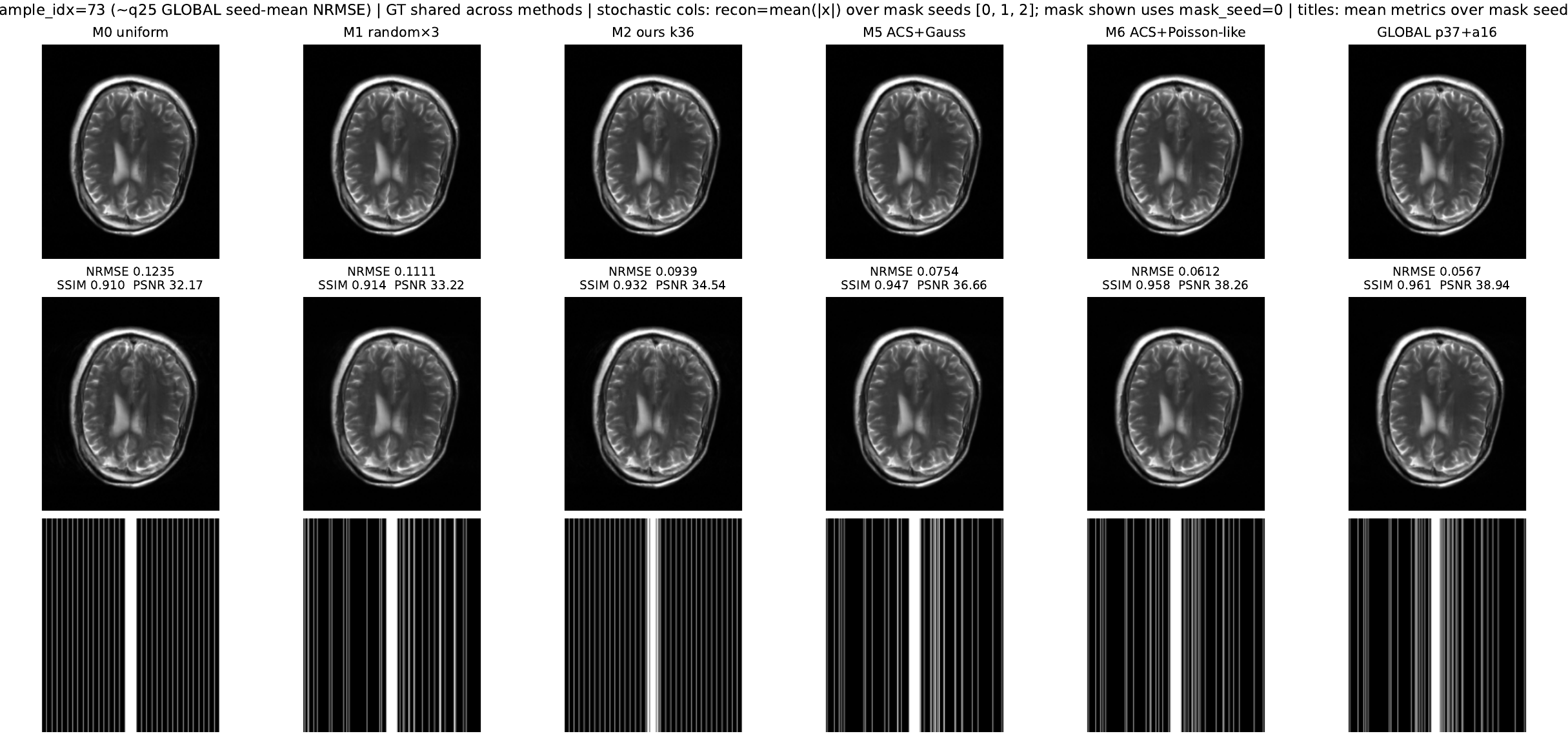}
\caption{Additional MR example, sample 073.}
\label{fig:appendix-mr-073}
\end{figure}

This subsection provides additional MR visualizations under the same $R=6$ Cartesian line-sampling protocol as Figure~\ref{fig:mr-sample}. Figure~\ref{fig:appendix-mr-037-copy} repeats the representative sample with the full original panel layout, and Figures~\ref{fig:appendix-mr-018}--\ref{fig:appendix-mr-073} show four additional samples used to check that the quantitative trends are not driven by a single visual example.

\end{document}